\documentclass[10pt,twocolumn,letterpaper]{article}
\usepackage{multirow} 
\usepackage[margin=1in]{geometry}  % Set margins
\usepackage{times}                 % Times font
\usepackage{graphicx}
\usepackage{amsmath}
\usepackage{amssymb}
\usepackage{booktabs}
\usepackage{bbm}
\usepackage[utf8]{inputenc}
\usepackage[T1]{fontenc}
\usepackage{url}
\usepackage{amsfonts}
\usepackage{nicefrac}
\usepackage{microtype}
\usepackage{xcolor}
\usepackage{algorithm}
\usepackage{algpseudocode}
\usepackage{float}
\usepackage{subcaption}
\usepackage{placeins}
\usepackage{xurl}
\usepackage[colorlinks=true, linkcolor=blue, citecolor=blue, urlcolor=blue, breaklinks=true]{hyperref}
\title{SurgXBench: Explainable Vision-Language Model Benchmark for Surgery}

\author{
Jiajun Cheng$^{1}$ \quad Xianwu Zhao$^{1}$ \quad Sainan Liu$^{2}$ \quad Xiaofan Yu$^{5}$ \quad Ravi Prakash$^{3}$ \\
Patrick J. Codd$^{3}$ \quad Jonathan Elliott Katz$^{4}$ \quad Shan Lin$^{1}$\\
$^{1}$Arizona State University, $^{2}$Intel Labs, \\
$^{3}$Duke University, $^{4}$University of Miami, $^{5}$University of California, Merced
}

\date{} % Remove date or add \date{\today}

\begin{document}
\maketitle

\begin{abstract}

Innovations in digital intelligence are transforming robotic surgery with more informed decision-making. Real-time awareness of surgical instrument presence and actions (e.g., cutting tissue) is essential for such systems. Yet, despite decades of research, most machine learning models for this task are trained on small datasets and still struggle to generalize. Recently, vision-Language Models (VLMs) have brought transformative advances in reasoning across visual and textual modalities. Their unprecedented generalization capabilities suggest great potential for advancing intelligent robotic surgery. However, surgical VLMs remain underexplored, and existing models show limited performance, highlighting the need for benchmark studies to assess their capabilities and limitations and to inform future development. To this end, we benchmark the zero-shot performance of several advanced VLMs on two public robotic-assisted laparoscopic datasets for instrument and action classification. Beyond standard evaluation, we integrate explainable AI to visualize VLM attention and uncover causal explanations behind their predictions. This provides a previously underexplored perspective in this field for evaluating the reliability of model predictions. We also propose several explainability analysis-based metrics to complement standard evaluations. Our analysis reveals that surgical VLMs, despite domain-specific training, often rely on weak contextual cues rather than clinically relevant visual evidence, highlighting the need for stronger visual and reasoning supervision in surgical applications. The code is provided in our public repository at: \url{https://jiajun344.github.io/SurgXBench-Explainable-Vision-Language-Model-Benchmark-for-Surgery-Project-Website/}.
\end{abstract}

\section{Introduction}
% P1: Bg & Challenges
Tremendous progress in robotic surgical systems has enabled high-precision, dexterous operations in minimally invasive settings, allowing access to narrow surgical spaces that are often unreachable via traditional open surgery \cite{ciuti2025robotic,bouget2017vision}. 
As robotic procedures expand, integrating digital intelligence can augment surgeons' capabilities through context-aware feedback and real-time decision support \cite{dupont2021decade,yip2025robot}. This integration enables more efficient procedures and more consistent, positive patient outcomes.
Despite strong motivation, and over two decades of research into developing AI technologies for critical tasks such as surgical instrument, tissue, and phase recognition, existing models are still lack accuracy and robustnees \cite{bouget2017vision, twinanda2016endonet, garcia2017toolnet, lin2020lc, ni2020attention, ross2021comparative, lin2021multi}. 
Thus, Current surgical robots remain fully controlled by human surgeons. This is largely due to the fact that most existing models are developed using small, manually annotated datasets and still struggle to consistently achieve good performance across visually diverse surgical data.

Recently, Vision-Language Models (VLMs), such as CLIP \cite{radford2021learning} and LLaVA \cite{liu2023visual}, have been trained by aligning visual and textual features rather than relying on curated labels. By learning from raw image-text pairs at scale, these models achieve strong generalization across many domains \cite{zhang2024vision}.
In surgical robotics, early studies have developed surgical VLMs using publicly available surgical lecture videos and educational materials, but their performance remains limited and not well understood \cite{yuan2023surgvlp,yuan2024hecvl,yuan2024peskavlp,rau2025systematic}.
% Recent efforts in surgical AI have begun to explore foundation models for surgical environments~\cite{schmidgall2024general}. SurgVLP, HecVL, and Peskavlp\cite{yuan2023surgvlp,yuan2024hecvl,yuan2024peskavlp} are trained using contrastive learning on surgical videos paired with text, with each method adding progressively more structure or supervision to improve alignment.
This indicates the need for systematic benchmark studies to analyze how existing VLMs, both general and surgical models, perform on surgical tasks. A more comprehensive understanding of their strengths and limitations is essential to guide the future development of surgical VLMs.

\begin{figure*}[t!]
\centering
\includegraphics[width=0.9\linewidth]{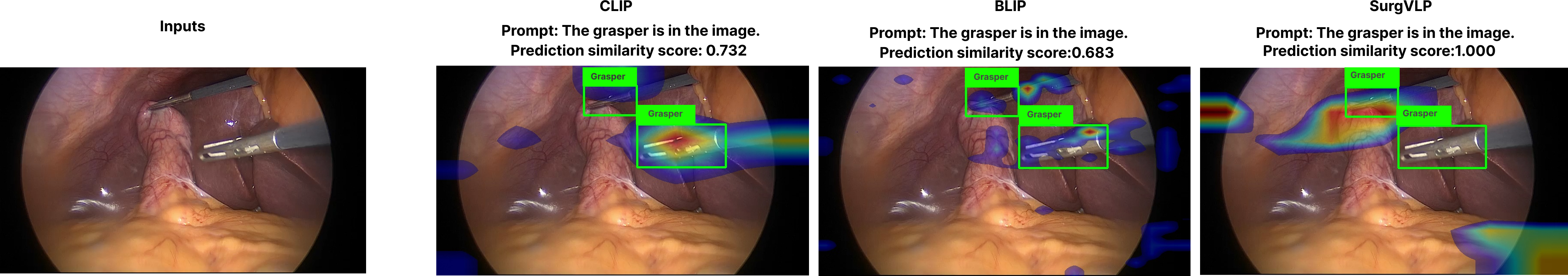}
\caption{
We evaluate two general VLMs, CLIP-ViT-B/32 \cite{radford2021learning}, BLIP \cite{li2022blip}, and a surgical VLM, SurgVLP \cite{yuan2023surgvlp}, on surgical instrument classification. The first image shows an input frame from the Cholec80BBox \cite{Jalal23b, Alshirbaji24} dataset. The next three images overlay Grad-CAM attention maps \cite{selvaraju2017gradcam,choe2020evaluating} on top of the original images, showing each model’s focus in response to the prompt \texttt{The grasper is in the image.}. The similarity scores above each image indicate the model’s prediction confidence (min-max normalized). Green bounding boxes and labels are ground truth instrument locations and types.}
\label{fig:intro_pic_examples}

\end{figure*}
% P3:  what's missing in current benchmark work: argue Rau's work -> XAI (either in this paragraph or next)
To the best of our knowledge, the only existing VLM benchmark in surgery is by by Rau \textit{et al.}~\cite{rau2025systematic}, which evaluates the zero-shot and few-shot performance of several state-of-the-art VLMs on multiple surgical vision tasks, including instrument and anatomy identification, action and phase recognition, and skill assessment. 
Their results also show that current surgical VLMs still perform poorly, sometimes even underperforming general VLMs on certain surgical tasks. Moreover, the performance of both general and surgical VLMs remains far below clinical requirements; for example, Rau \textit{et al.} report precision below 60\% for instrument recognition in both zero-shot and 5-shot settings. 
While Rau \textit{et al.}’s benchmark covers many tasks, it only reports standard evaluation metrics, which primarily reflect the average poor performance across datasets, but it lacks deeper insight into why these models fail and which aspects require improvement to better adapt VLMs for surgical applications in the future.

% P4: Our work: explainability -> contributions
To this end, our work goes beyond standard performance reporting. We integrate widely used explainable AI (XAI) techniques \cite{zablocki2022explainability}, including Grad-CAM \cite{selvaraju2017gradcam,choe2020evaluating} and CLEANN \cite{rohekar2023causal,stan2024lvlm}, into the evaluation of VLMs to visualize their attention and causal reasoning behind their predictions. Then, whether these models’ reasoning processes align with clinical relevant cues, such as the instrument locations and the tissues they interact with, is analyzed. 
Figure \ref{fig:intro_pic_examples} shows a representative example from our explainability analysis: the surgical model SurgVLP \cite{yuan2023surgvlp} correctly classifies the instrument as a grasper with higher confidence (\textit{i.e.}, higher similarity score), but compared to general VLMs such as CLIP \cite{radford2021learning} and BLIP \cite{li2022blip}, the attention of SurgVLP does not align well with the most relevant regions, which are the instrument regions in this example. This raises concerns about the reliability of the surgical VLM's predictions, which are not adequately reflected by standard evaluation metrics. 
To systematically assess these observations, we further propose several evaluation metrics to quantitatively assess how well the models’ attention and decision-making processes align with key clinical cues.
As the first step toward explainability-based benchmarking for surgical tasks, we focus on two fundamental and critical tasks, surgical instrument and action classification. They serve as the foundation for many high-level applications such as surgical phase recognition and skill assessment. Moreover, our explainability-driven analysis pipeline and proposed evaluation metrics can be flexibly extended to other surgical tasks, as long as relevant clinical cues are defined with appropriate domain knowledge. In summary, our main contributions are as follows:
\begin{itemize}
    \item We introduce SurgXBench, the first VLM benchmarking framework for surgical tasks that integrates explainability. By analyzing model attention and causal reasoning, SurgXBench reveals a significant mismatch between where surgical VLMs focus and clinically relevant regions, even for correct predictions. This indicates the need for stronger visual grounding and semantic supervision in future models.
    \item We propose several new metrics to quantitatively assess attention alignment. These metrics extend our qualitative findings based on the explainability analysis and complement standard task metrics, enabling a more comprehensive evaluation of model prediction reliability. 
    \item For instrument action classification, we leverage an off-the-shelf optical flow method with camera motion correction to automatically extract instrument-tissue interaction regions. This augments existing datasets with ground truth for where critical instrument interactions occur and enables quantitative evaluation of model attention for surgical actions, an aspect not rigorously studied before.
    % \item\jc{ We explore potential future development of VLMs by proposing hypotheses about potential factors contributing to current VLM limitations in surgical applications.}
\end{itemize}
\section{Related Work}

\textbf{General and Surgical VLMs:} VLMs have emerged as powerful frameworks for multimodal understanding and generally falling into two categories: contrastive VLMs and Large Visual language models (LVLMs). Contrastive models like CLIP~\cite{radford2021learning} and BLIP~\cite{li2022blip} learn to align visual and textual features, while LVLM like LLaVA~\cite{liu2023visual} integrate vision encoders with large language models to directly generate text responses with more complex reasoning.
General VLMs have demonstrated strong generalization in many fields of machine intelligence, vision, and robotics, largely due to their ability to learn from massive amounts of raw multimodal data available online. However, without training on surgical data, they lack domain-specific knowledge needed to understand surgical environments. To address this gap, 
surgical VLMs have recently been developed to address the unique challenges of surgical settings. SurgVLP~\cite{yuan2023surgvlp} pioneered large-scale surgical pretraining using surgical video lectures with CLIP-style contrastive learning to align video clips and textual descriptions. HecVL~\cite{yuan2024hecvl} enhanced this approach by introducing hierarchical supervision by pairing videos with both free-form text and structured phase-level annotations. PeskaVLP~\cite{yuan2024peskavlp} further advanced this line of work by incorporating LLM-augmented surgical knowledge and applying procedure-aware loss to improve cross-modal temporal alignment. 
% In this work, we select several widely used general and surgical VLMs for benchmarking.
Yet, current surgical VLMs still exhibit limited performance, often offering only marginal advantages over general VLMs and sometimes even underperforming, as also confirmed by the only prior benchmark study in this domain by Rau \textit{et al.} \cite{rau2025systematic}. This indicates the need for a deeper understanding of the limitations of existing VLMs to guide the development of more effective surgical models. Since Rau \textit{et al.} focused solely on standard benchmarking based on traditional task metrics, we propose to complement such evaluation with explainable AI technologies to analyze the reasoning behind VLM predictions.

%Most prior work with VLMs in surgical tasks has focused on temporal modeling and phase recognition, whereas we emphasize the spatial and semantic grounding abilities of VLMs in zero-shot settings.

\textbf{Explainable AI (XAI):}
While deep learning models achieve superior performance, their lack of interpretability poses challenges. Several visualization methods have been developed to identify influential image regions and reveal the evidence behind model predictions, providing insights into model explainability. Visualization methods like Grad-CAM~\cite{selvaraju2017gradcam} highlight important image regions by computing gradients with respect to feature maps, with extensions including attention-based approaches~\cite{chefer2021generic}, Grad-CAM++~\cite{chattopadhay2018gradcampp} for multiple objects, and Score-CAM~\cite{wang2020scorecam} using output scores directly. Additionally, causal interpretation methods like CLEANN~\cite{rohekar2023causal,stan2024lvlm} provide explanations by learning causal structures from attention mechanisms, testing conditional independence between token pairs to identify which input tokens would change model output if masked. However, limited research in medical XAI has focused on surgical contexts, with most lacking proper explainability evaluation. Zhang et al.~\cite{zhang2022applications} noted the absence of unified XAI evaluation methods and standardized metrics for trustworthy AI on surgical tasks like phase and instruments recognition, highlighting the critical need for clinically meaningful explainability frameworks. 

\section{Proposed Methods}
In this section, we introduce the benchmarked surgical tasks in Section~\ref{sec:task}, explain the involved explainable AI~\cite{zablocki2022explainability} methods in Section~\ref{sec:attn_vis}, outline the questions guiding our explainability analysis in Section \ref{sec:prob_def}, and present the proposed explainability-based metrics in Section~\ref{sec:metric}.

\subsection{Task Definition} \label{sec:task}

Our benchmark focuses on two fundamental and important tasks for intelligent robotic surgery: (i) \emph{instrument classification}, and (ii) \emph{triplet recognition}. 

\textbf{Instrument Classification:} The goal of this task is to determine which surgical instruments are present in a given image. Awareness of the presence of instruments will enhance the control and use of them \cite{bouget2017vision}. Moreover, since instrument usage is closely related to surgical phases, accurate instrument classification also supports downstream applications such as phase recognition and skill assessment~\cite{twinanda2016endonet}.

\textbf{Triplet Recognition:}
In addition to instrument classification, triplet recognition is a task aims to also predict the associated instrument action as subject-verb-object triplets~\cite{nwoye2022rendezvous}, such as \textit{(scissors, cut, tissue)} triplet, describing which instrument performs what action on which target. This usage information can support downstream needs such as surgical workflow analysis and identification of potential procedure risks.

\subsection{XAI Visualization Techniques} \label{sec:attn_vis}
We employ three XAI methods to identify important regions for predictions. For contrastive models with ResNet backbone~\cite{he2016deep}, Grad-CAM computes gradients of similarity scores with respect to features to produce spatial attention heatmaps. For LVLMs, Grad-CAM computes gradients of token logits with respect to features~\cite{zhang2024redundancy}. For transformer architectures~\cite{dosovitskiy2020image}, we use gradient-weighted attention rollout~\cite{chefer2021generic}, which leverages self-attention matrices to identify contributing patches through relevance propagation. Additionally, CLEANN~\cite{rohekar2023causal,stan2024lvlm} analyzes causal relationships between token pairs, discovering how visual attention causally influences language generation. Mathematical formulations are in the supplementary section A.

\subsection{Problem Definition} \label{sec:prob_def}
A main objective of our work is to evaluate VLMs from a novel perspective using XAI technologies to assess whether the models’ prediction processes align with clinically relevant cues. Specifically, our explanability-based analysis focuses on the following questions: 
% Building on the use of Grad-CAM and CLEANN as tools to visualize model attention, we aim not only to  benchmark the performance of VLMs on surgical data, a main objective of this work is to assess whether their predictions are grounded in the clinical-relevant cues. Formally, this motivates the following question:
\begin{itemize}
\item \textbf{(A) When the model makes a correct prediction:}
\begin{itemize}
\item (i) Does its attention or causal reasoning behind its prediction focus on clinically relevant visual cues?
\item (ii) Are there cases where the model predicts correctly without attending to the relevant image regions, suggesting that the prediction may be unreliable or coincidental?
\end{itemize}

\item \textbf{(B) When the model makes an incorrect prediction:}
\begin{itemize}
\item (i) Is the incorrect prediction due to the model attending to irrelevant visual cues?
\item (ii) Does the model focus on relevant cues, but unable to recognize the instrument type due to lack of domain knowledge?
\end{itemize}
\end{itemize}
\subsection{Explainability-based Metrics} \label{sec:metric}
To quantify the explainability-based analysis, as inspired by \cite{selvaraju2017gradcam} and \cite{choe2020evaluating}, we propose several metrics to measure the alignment between the visual cues utilized by the models and clinically relevant cues, thereby helping answer questions outlined in Section \ref{sec:prob_def}. These explainability-based metrics are reported alongside standard evaluation metrics for involved tasks to provide a more comprehensive evaluation of the model.
% % For benchmarking VLMs on surgical understanding, 
% we evaluate quantitative performance through explainability (XAI) metrics that provide complementary insights into model behavior. Drawing inspiration from the Grad-CAM method~\cite{selvaraju2017gradcam} and \cite{choe2020evaluating}, 
% % an evaluation method that involves measuring whether AI models attend to the correct object regions rather than just achieving high classification accuracy, 
% we propose attention-based metrics since semantically meaningful attention regions produce more interpretable predictions.
% %For instrument classification, we propose Attention Alignment (AA) and Attention Coverage (AC) metrics, measuring whether model attention aligns with annotated instrument regions. For triplet recognition, we propose a metric leveraging optical flow with camera motion compensation to assess whether attention aligns with action-relevant dynamic regions. 
% We also adopt standard evaluation metrics. 
% % Instrument classification uses precision, recall, and F1-score computed by thresholding image–text similarity score. Triplet recognition follows the evaluation framework from~\cite{nwoye2022rendezvous}, evaluating instrument–verb, instrument–target, and full triplet prediction accuracy. 
Detailed definitions for standard metrics are provided in supplementary material section D.

\subsubsection{Metrics for Instrument Classification: AA, AC} \label{sec:instrument metrics}
Let $I$ denote an input image of size $H \times W$. Grad-CAM is used to produce a heatmap $L\in \mathbb{R}^{H \times W}$
% using techniques from \ref{sec:attn_vis}, 
% let $L(i, j) 
, in which each pixel reflects the attention value of that location for the model's prediction. We then apply a percentile-based threshold $\tau$ to retain information about the relative intensity of attention across spatial locations. We pick top $\tau$ percent of attention value in the heatmap $L$. The predicted attention region is then defined as:
\begin{equation}
\mathcal{A}_\tau = \left\{ (i, j) \mid L(i,j) \geq \mathrm{Quantile}_{1 - \tau}(L) \right\}.
\end{equation}
where $L(i,j)$ denotes the pixel at $i$-th row and $j$-th column in $L$. 

Let $\mathcal{C} = \{ c_1, \dots, c_K \}$ denote the set of ground-truth class labels (e.g., surgical instruments) present in the image, and $N_k$ is the number of appearnces of class $c_k$ in the image,For each class $c_k \in \mathcal{C}$, let $\mathcal{G}_{c_k}^{(r)} \subseteq \{1,\dots,H\} \times \{1,\dots,W\}$ denote the set of pixels in the $r^\text{th}$ annotated region of class $c_k$, for $r = 1,\dots,N_k$. Let $\hat{c}$ be the class predicted by the model. We define the set of all annotated pixels (regardless of class) as:
\[
\mathcal{G}_{\text{all}} = \bigcup_{k=1}^{K} \bigcup_{r=1}^{N_k} \mathcal{G}_{c_k}^{(r)} \subseteq \{1,\dots,H\} \times \{1,\dots,W\}.
\]
We define the set of pixels corresponding to the predicted class $\hat{c}$ as:
\[
\mathcal{G}_{\hat{c}} =
\begin{cases}
\displaystyle \bigcup_{\substack{\\ c_k = \hat{c}}} \bigcup_{r=1}^{N_k} \mathcal{G}_{c_k}^{(r)}, & \text{if } \hat{c} \in \mathcal{C} \\
\emptyset, & \text{otherwise}
\end{cases}
\]

\textbf{Thresholded Attention Coverage (AC):}
This metric measures how much of the model's attention region overlaps with any annotated object, regardless of class correctness:
\begin{equation}\label{eq:AC}
\text{AC} = \frac{|\mathcal{A}_\tau \cap \mathcal{G}_{\text{all}}|}{|\mathcal{A}_\tau|}.
\end{equation}
%It reflects the fraction of model attention that lands inside any annotated region. This is a less strict metric that serve as a soft proxy to access whether the model is at least focusing on some semantically meaningful area. 

\textbf{ Thresholded Attention Alignment (AA):}
This metric evaluates whether the model's attention aligns specifically with the annotated regions of the predicted class $\hat{c}$. If $\hat{c} \notin \mathcal{C}$, the score is defined to be zero:
\begin{equation}\label{eq:AA}
\text{AA} =
\begin{cases}
\displaystyle \frac{|\mathcal{A}_\tau \cap \mathcal{G}_{\hat{c}}|}{|\mathcal{A}_\tau|}, & \text{if } \hat{c} \in \mathcal{C} \\
0, & \text{otherwise}
\end{cases}
\end{equation}

\subsubsection{Metrics for Triplet Recognition: RAS} When it comes to localizing action spatial regions, things get more complex. While actions can be learned by looking at instruments \cite{twinanda2022endonet}, real applications involve tool-tissue interactions where deformable tissues carry meaningful information. Motion reflects rich regions of interest across consecutive frames—information not captured by standard semantic segmentation datasets.

\begin{figure}[t!]
  \centering
  \includegraphics[width=0.3\textwidth]{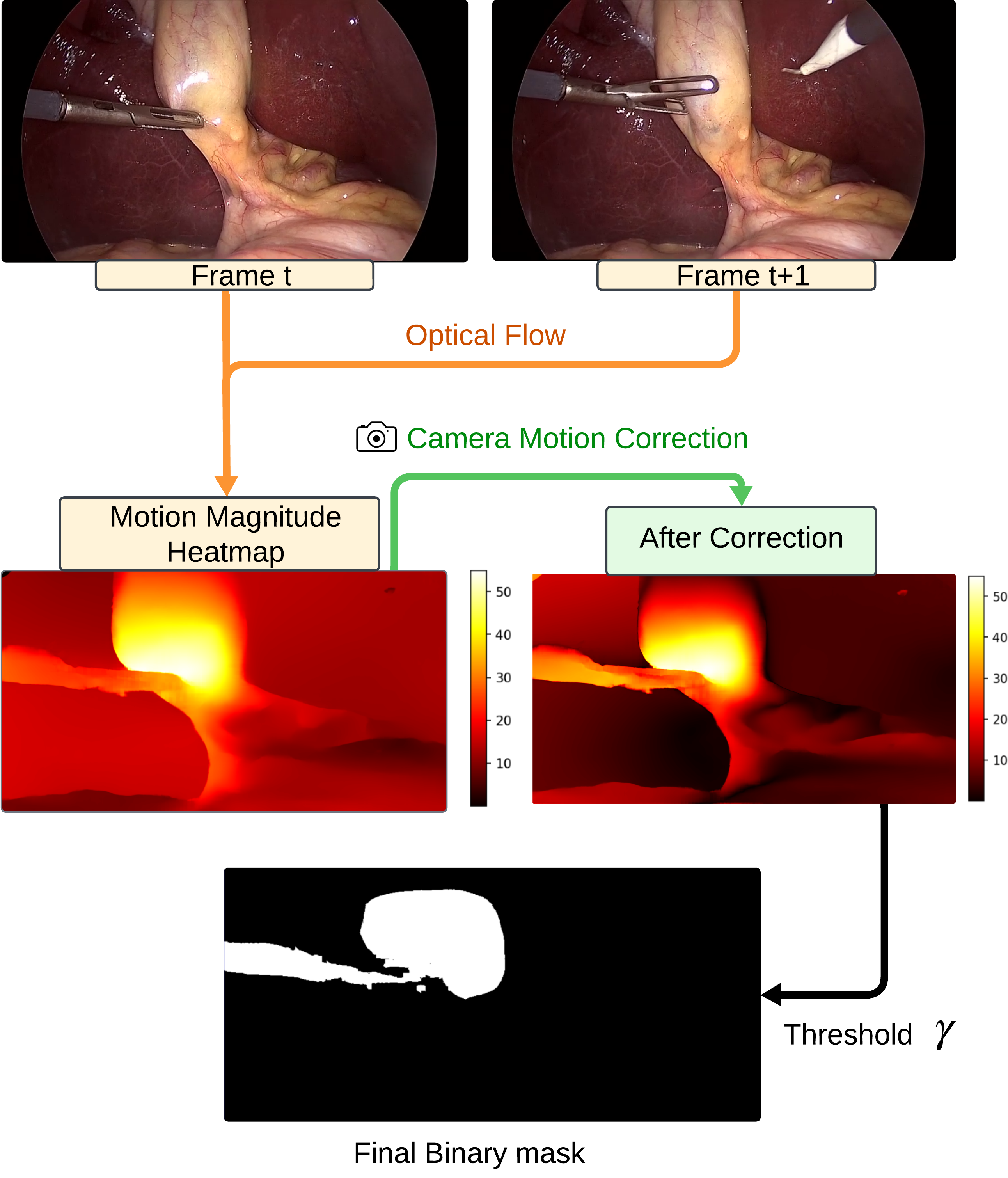}
\caption{Pipeline of extracting instrument-tissue interaction region for benchmark triplet recognition task. Optical flow is estimated by RAFT~\cite{teed2020raft}, and heatmaps represent motion magnitude computed from the flow vectors. Camera correction is applied to reduce background motion in the heatmap. Threshold $\gamma$ retains high motion magnitude regions to generate binary masks.}
\label{fig:raft pipline}
\end{figure}

\paragraph{RAFT and Camera Motion Correction:} To address the lack of instrument-tissue interaction reigion annotation, we employed an off-the-shelf optical flow method RAFT \cite{teed2020raft} to estimate optical flow between consecutive frames, computing dense pixel-wise motion vectors to generate masks highlighting high-motion areas. To reduce the effects of camera movement, we model camera motion as a linear combination of Pan, Tilt, Zoom, and Roll operations \cite{almeida2009robust}, using bidirectional RAFT flow with least-squares optimization to subtract camera motion and isolate true interaction motion. Applying percentile threshold $\gamma$ to the corrected RAFT heatmap yields binary masks marking high-motion regions (Figure~\ref{fig:raft pipline}).

\paragraph{RAFT Attention Score (RAS):} Based on this motion-aware supervision, where $\mathcal{A}_{\tau^v}$ denotes thresholded attention heatmap of a triplet prediction, and $\mathcal{M}_{\text{raft}} \in \{0,1\}^{H \times W}$ denotes the binary mask generated by RAFT method, $\mathcal{G}_{s}$ the predicted instrument pixel set, and $\mathcal{G}_{\text{all}}$ the union of all ground-truth instrument regions.

\paragraph{Local RAS:} Measures alignment of verb attention with motion or region of the predicted instrument:
\begin{equation}\label{eq:ras-local}
\text{RAS}_{\text{local}} = \frac{|\mathcal{A}_{\tau^v} \cap (\mathcal{M}_{\text{raft}} \cup \mathcal{G}_{s})|}{|\mathcal{A}_{\tau^v}|}.
\end{equation}

\paragraph{Global RAS:} Measures alignment with all ground-truth instrument regions:
\begin{equation}\label{eq:ras-global}
\text{RAS}_{\text{global}} = \frac{|\mathcal{A}_{\tau^v} \cap (\mathcal{M}_{\text{raft}} \cup \mathcal{G}_{\text{all}})|}{|\mathcal{A}_{\tau^v}|}.
\end{equation}
This score quantifies how much attention of triplet predictions overlap with instrument-tissue interaction regions.

\section{Experiments Set up}
\label{sec:experiments}
\subsection{Dataset}
We use the following two publicly available robotic-assisted laparoscopic datasets. 

% derived from the Cholec80 dataset~\cite{Twinanda16}. 
\textbf{Cholec80BBox}~\cite{Jalal23b, Alshirbaji24} includes 5 videos with 51.7K frames annotated with instrument bounding boxes and class labels. We use this dataset to benchmark instrument classification, with class labels as ground truth for instrument classification evaluation and bounding boxes as ground truth for computing explainability-based metrics. 

\textbf{CholecT45}~\cite{nwoye2022rendezvous} consists of 45 videos with 90.5K frames annotated with frame-level triplet annotations. This dataset is used to benchmark triplet recognition. Since our $\text{RAS}_{\text{local}}$ metric \eqref{eq:ras-local} requires instrument bounding box annotations to measure attention alignment, we use only 5,957 frames from videos \#42 and \#43, which are the only ones with such annotations.
% because our RAFT-generated ground truth masks lack clear instrument type labels, and these two videos uniquely contain both triplet annotations and instrument bounding boxes from Cholec80BBox, enabling spatial alignment evaluation.

\begin{table}[t]
\centering
\scalebox{0.8}{ 
\small
\begin{tabular}{lcc}
\toprule
\textbf{Model} & \textbf{Type} & \textbf{Training Data} \\
\midrule
SurgVLP & Contrastive & Surgical Video \\
HecVLP & Contrastive & Surgical Video \\
PeskaVLP & Contrastive & Surgical Video \\
CLIP & Contrastive & Web Image-Text \\
BLIP & Contrastive & Web Image-Text \\
LLaVA-v1.5-7B$^\ddagger$ & LVLM & Web Image-Text  \\
\bottomrule
\end{tabular}
}
\caption{Models used in this study. LLaVA-v1.5-7B$^\ddagger$ uses LLaMA2 as LLM backbone.}
\label{tab:overalldatamodel}
\end{table}

\begin{figure*}[t!]
    \centering
    \includegraphics[width=\textwidth]{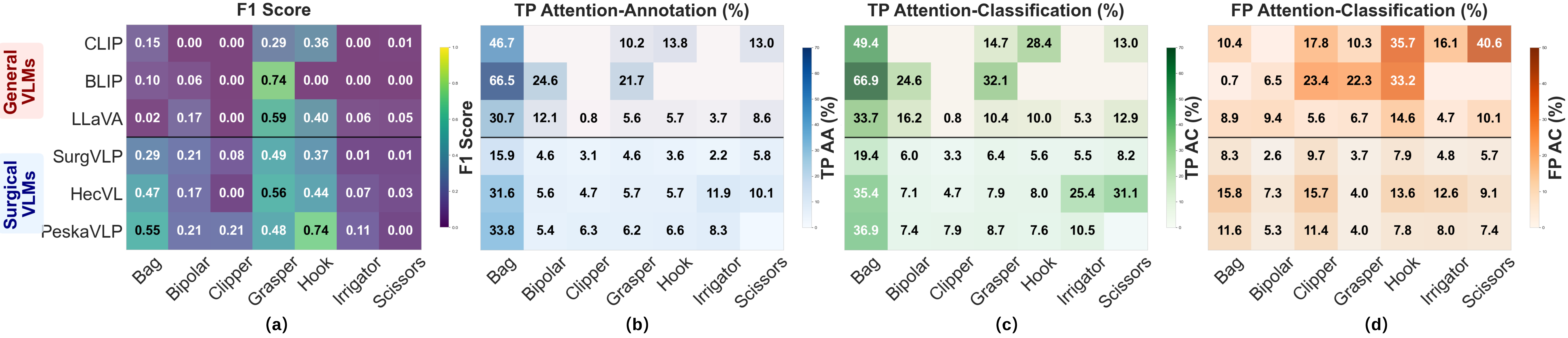}
\caption{Instrument classification results of General VLMs (CLIP\cite{radford2021learning}, BLIP\cite{li2022blip}, LLaVA\cite{liu2023visual}) and Surgical VLMs (SurgVLP\cite{yuan2023surgvlp}, HecVL\cite{yuan2024hecvl}, PeskaVLP\cite{yuan2024peskavlp}) on Cholec80BBox dataset. Results are reported using standard metric (a) F1 scores, and proposed explainability-based metrics that measure model attention alignment with clinically relevant cues: (b) TP\_AA, (c) TP\_AC, and (d) FP\_AC.}\label{fig: AA_AC_results}
\end{figure*}
\begin{figure*}[t!]
    \centering
    \includegraphics[width=0.9\textwidth]{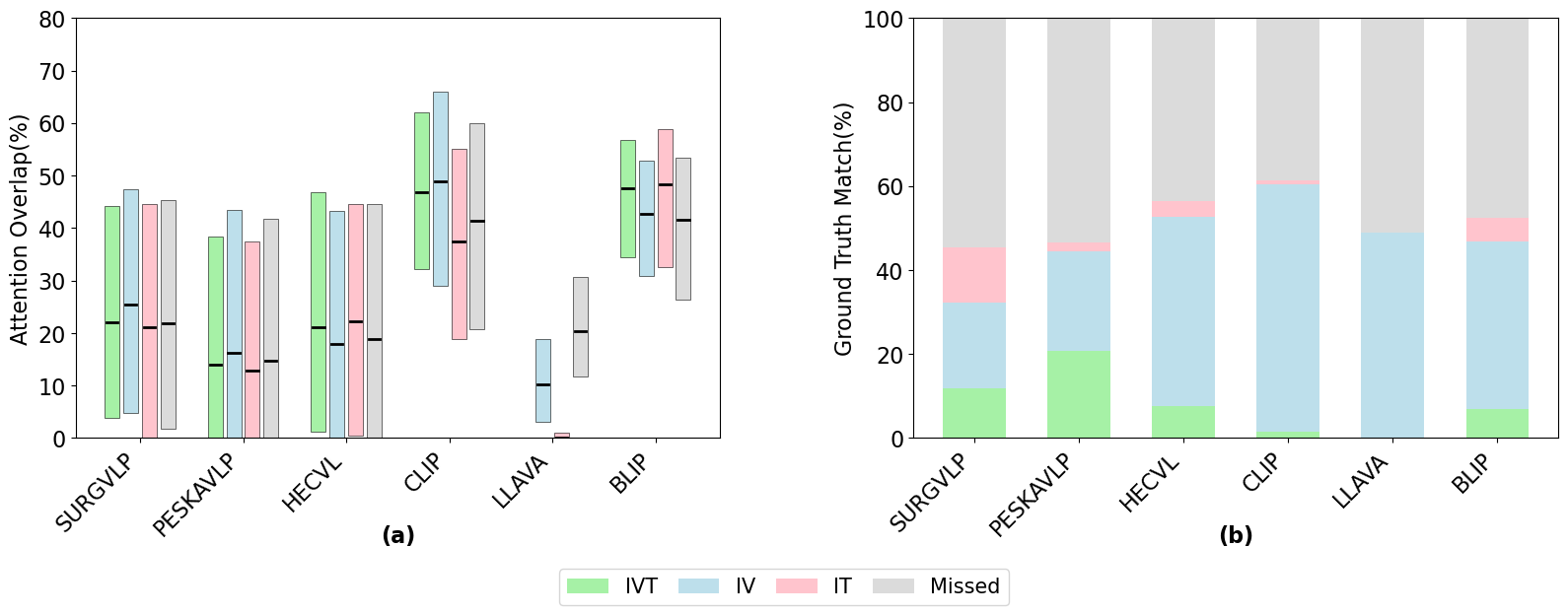}
    \caption{Triplet recognition results on CholecT45\cite{nwoye2022rendezvous}dataset. (a) Box plots of RAS\textsubscript{local} for IVT, IV, and IT matches, and RAS\textsubscript{global} for missed cases. (b) Percentage breakdown of how ground truth triplets are matched by model predictions: IVT (complete match), IV (instrument-verb match), IT (instrument-target match), and Missed (no instrument match).}\label{fig:triplet_result}
\end{figure*}
\subsection{Benchmarked Models}
We evaluate the zero-shot performance of six pre-trained VLMs, 
% categorized into two groups to assess general versus surgical VLMs performance
including three \textbf{General VLMs} CLIP, BLIP, LLaVA, along with 
% are trained on large-scale natural image-text pairs and have strong zero-shot capabilities across various domains.
three \textbf{Surgical VLMs} SurgVLP, HecVL, and PeskaVLP. 
%are all contrastive learning models trained on surgical video datasets to understand visual-linguistic information in surgical environments. 
 More details of the benchmarked models are provided in Table~\ref{tab:overalldatamodel}.

%CLIP uses contrastive learning, BLIP employs multimodal encoder-decoder architecture, and LLaVA integrates vision encoders with language models through visual instruction tuning. \textbf{Surgical VLMs} are specifically trained on surgical video datasets to understand visual-linguistic information in surgical environments. All three models use contrastive learning approaches tailored for surgical contexts. Details are in Table~\ref{tab:overalldatamodel}.

\subsection{Evaluation Strategy}
We report standard classification metrics (precision, recall, F1) \cite{powers2020evaluation} and our attention alignment scores to assess prediction trustworthiness beyond accuracy. To address question \textbf{(A-ii)} from Section~\ref{sec:prob_def}, we design \textbf{Attention-Based Occlusion Analysis} inspired by counterfactual explanations~\cite{goyal2019counterfactual}. Contrastive models compute image-prompt similarity; prompts exceeding threshold $\delta$ form predicted set $\hat{\mathcal{P}}$. We generate heatmaps using post-hoc XAI methods (Section~\ref{sec:attn_vis}) for each predicted class $\hat{p} \in \hat{\mathcal{P}}$. Details of implementation are in supplementary section C.

%\subsubsection{Prompt Construction and Prediction Generation} \label{sec:prompt_design}
%To align with VLMs' prompt-based input format, we construct template-based prompts. For instrument classification, we use: \textit{``an image showing a \{instrument class\} in use''}. For triplet reasoning, we use: \textit{``I use a \{instrument\} to \{verb\} the \{target\}''}. For CLIP, SurgVLP, HecVL, and PeskaVLP, each model computes cosine similarity between the image and all prompt candidates. Prompts above a predefined percentile threshold $\delta$ form the predicted set $\hat{\mathcal{P}}$. BLIP uses Image-Text Matching (ITM) scores for binary classification. For LLaVA, we design task-specific queries. For instrument classification: \textit{``List all surgical instruments you see. Options: \{grasper, hook, scissors, clipper, irrigator, bipolar, null\}''}. For triplets: \textit{``Find all instruments and their actions on tissues''} with JSON output format \textit{\{"instrument": [...], "verb": [...], "target": [...]\}} and instrument-action compatibility constraints. Detilaed prompts and implementation are provied in supplementary material. We generate heatmaps using post-hoc XAI techniques from Section~\ref{sec:attn_vis} for each predicted class $\hat{p} \in \hat{\mathcal{P}}$ and evaluate spatial alignment as follows:

%To assess whether model predictions are supported by relevant visual evidence and to address the questions outlined in Section~\ref{sec:prob_def}, we apply the following attention analysis to each image in the \textbf{Cholec80BBox} and \textbf{CholecT45} dataset.

\subsubsection{Instrument Classification Task}  
For each image, Instrument predictions are categorized as True Positive (TP) or False Positive (FP), and we compute alignment scores using post-hoc explainability methods (Section~\ref{sec:attn_vis}). Let $\mathcal{P}$ denote the ground truth set for an image. For TP predictions ($\hat{p} \in \mathcal{P}$), both explainability-based metrics for instrument classification task, AA~\eqref{eq:AA} and AC~\eqref{eq:AC} scores are computed, while for FP predictions ($\hat{p} \notin \mathcal{P}$), only AC~\eqref{eq:AC} is calculated. We report instrument-wise F1-scores on the \textbf{Cholec80BBox} dataset, along with True Positive AA and AC scores and False Positive AC scores. This enables statistical analysis of attention alignment patterns, addressing questions \textbf{A-i}, \textbf{B-i}, and \textbf{B-ii} from Section~\ref{sec:prob_def}.

\subsubsection{Triplet Recongition Task}  
We apply a percentile threshold \(\gamma\) to RAFT motion heatmaps to obtain binary masks and report attention alignment via RAS score box plots on videos 42 and 43 from the \textbf{CholecT45} dataset. We evaluate each ground truth by combining standard triplet metrics \cite{nwoye2022rendezvous} with our explainability-based metrics as follows:
\begin{itemize}
    \item \textbf{IVT match:} All components (instrument, verb, target) match in a top-$k$ prediction. We compute $\text{RAS}_{\text{local}}$.

    \item \textbf{IV match:} If no IVT match, but a top-$k$ prediction matches instrument and verb, we compute $\text{RAS}_{\text{local}}$.

    \item \textbf{IT match:} If no IV or IVT match, but a top-$k$ prediction matches instrument and target, we compute $\text{RAS}_{\text{local}}$.

    \item \textbf{Missed:} If no top-$k$ prediction matches the instrument, we compute average $\text{RAS}_{\text{global}}$ over all predictions.
\end{itemize}

 This captures the relationship between prediction correctness and attention behavior, addressing \textbf{A-i}, \textbf{B-i}, and \textbf{B-ii} in Section~\ref{sec:prob_def}. For LLaVA, we apply the same evaluation using all generated prediction set.

\begin{figure*}[t!]
    \centering
    \includegraphics[width=\textwidth]{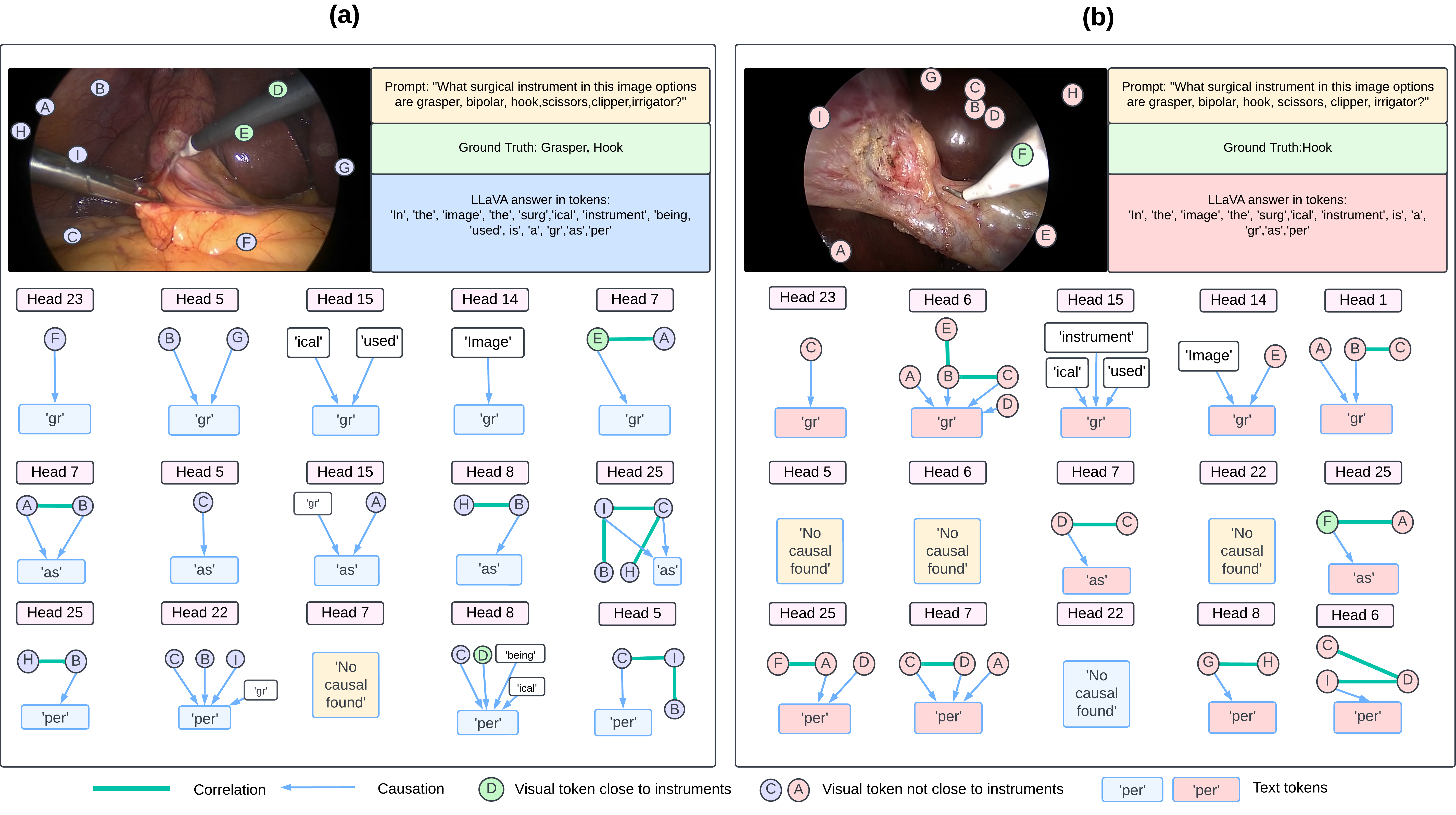}
\caption{
   LLaVA predictions with CLEANN-generated causal graphs (radius 3, top-5 attention heads). (a): Correct \emph{grasper} identification; (b): \emph{hook} misclassified as \emph{grasper}.
}\label{fig:causal}
\end{figure*}

\textbf{Attention-Based Occlusion Analysis:}\label{sec:occlusion exp} To evaluate question \textbf{A-ii}, we use VID42 and VID43 from CholecT45, selecting top-5 prompts as predictions. For frames with IVT-covered triplets, we apply Grad-CAM to generate heatmaps, retain top $1-\tau$ high-attention regions using percentile threshold $\tau$, then occlude these regions with black pixels following~\cite{zeiler2014visualizing}. We report effects on IVT coverage rates and similarity scores.

\textbf{Causal Graph Analysis:} \label{sec:causal_graph}
To visualize how the model attributes predictions to visual input, we apply CLEANN to generate causal graphs from the top five attention heads which is a parallel mechanisms where each head in the transformer attention learns to focus on different visual features and relationships that attend most strongly to visual tokens during the generation of each prediction token.

%\paragraph{Cholec80 BBox:}
%We use the \textbf{Cholec80 BBox} dataset~\cite{Jalal23b, Alshirbaji24}, an extension of Cholec80~\cite{Twinanda16}, which provides bounding box annotations for surgical instruments across 80 cholecystectomy videos. This dataset is used for the \textbf{instrument classification task}, allowing us to evaluate model predictions using spatially grounded metrics such as \textbf{$\tau$-AC}~\eqref{eq:AC} and \textbf{$\tau$-AA}~\eqref{eq:AA}. It is also used as a source of spatial supervision when computing attention alignment for triplet classification in CholecT45.
%\paragraph{CoPESD:}
%We also include the \textbf{CoPESD} dataset~\cite{Wang24}, the first multimodal dataset for endoscopic submucosal dissection (ESD), which contains over 17,000 images with fine-grained, multi-level instrument annotations. This dataset is used for evaluating \textbf{instrument classification} and enables spatial reasoning analysis via the \textbf{$\tau$-AC} and \textbf{$\tau$-AA} metrics.
%\paragraph{CholecT45:}
%We use the \textbf{CholecT45} dataset~\cite{Yuan24b}, which consists of 45 laparoscopic surgery videos annotated with action triplets of the form $\langle \texttt{instrument}, \texttt{verb}, \texttt{target} \rangle$. This dataset supports the \textbf{triplet classification task}, where the model selects the most semantically aligned structured action description. For spatial evaluation, we pair CholecT45 with bounding box annotations from Cholec80 BBox.

 \begin{figure*}[t!]
    \centering
    \includegraphics[width=\textwidth]{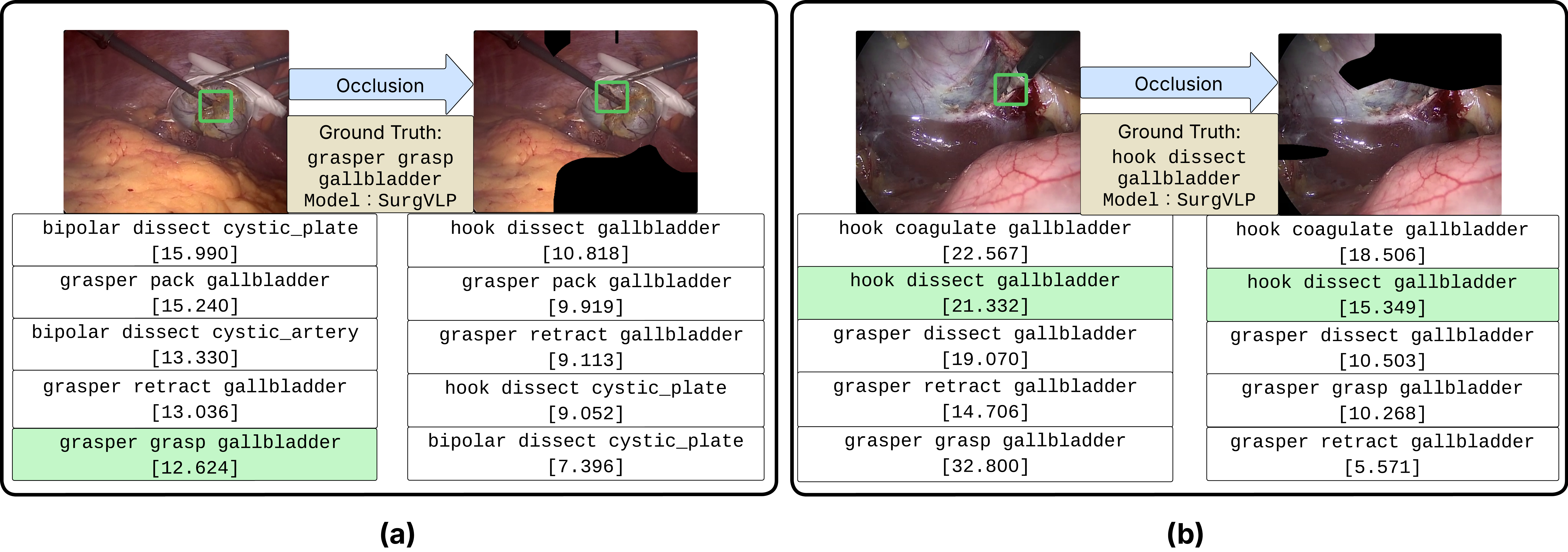}
\caption{
Counterfactual occlusion analysis results on SurgVLP. We picked two frames with the two most common instruments in the CholecT45 dataset as examples. Prompts with top-5 predicted similarity scores are shown, with the instrument tip marked by a green box in the image and corresponding prompt marked with a green background indicating IVT coverage. 
(a) An example of occluding irrelevant areas reducing the similarity score of the correct prediction, removing it from top-5 and leading to a wrong prediction. (b) An example of occluding relevant regions, in contrast, does not affect the prediction with correct prediction still has a relatively high similarity score. %Green prompts indicate IVT coverage; Green boxes mark instrument areas.
}
\label{fig:conterfactual visual explaination}
\end{figure*}

\section{Results}\label{sec: results}

%In this section, we present the results of the three previously mentioned experiments, beginning with attention quality metrics-based quantitative analysis on the instrument classification which summarized in Figure~ \ref{fig: AA_AC_results} and triplet recognition in Figure~\ref{fig:triplet_result}. We then report the results of the attention-based occlusion analysis which summarized in Figure~\ref{fig:conterfactual visual explaination}. Finally, to further explore the explainability of the large vision-language model LLaVA, we present findings from the Causal Graph Analysis, which leverages causal probabilistic graphs.

\subsection{Instruments Attention Quality Analysis}
Figure~\ref{fig: AA_AC_results} shows instrument classification results evaluated using both standard and proposed explainability-based metrics. General contrastive learning VLMs outperform surgical VLMs in attention alignment with higher AA and AC scores across most instruments regardless of prediction correctness, with BLIP showing strongest attention alignment (\textit{bag} TP\_AC: 66.9\%, \textit{grasper} TP\_AC: 32.1\%). This suggests general contrastive learning VLMs can localize instrument regions but misclassify them due to lacking surgical knowledge, while surgical VLMs exhibit better classification performance on most of instruments, with PeskaVLP achieves highest F1 for multiple classes (\textit{hook}: 0.7410, \textit{bipolar}: 0.7899, \textit{irrigator}: 0.7962) despite lower attention scores (\textit{hook} TP\_AA: 6.6\%, TP\_AC: 7.6\%). This suggests that surgical VLMs may rely on visual shortcuts rather than relevant cues, we further investigate through occlusion studies in Section \ref{sec:result_occlusion_study}.
% demonstrating decoupling between attention alignment and classification performance examined in our occlusion study. 
Additional detailed boxplots and precision recall results are provided in supplementary material section E.1.

\subsection{Triplets Attention Quality Analysis}
Figure \ref{fig:triplet_result} shows triplet recognition results. In Figure~\ref{fig:triplet_result}(a), contrastive general VLMs exhibit higher $\text{RAS}_{\text{local}}$ and $\text{RAS}_{\text{global}}$ scores (BLIP: 47.5\%) than surgical VLMs (SurgVLP: 22.1\%), while LLaVA shows weaker attention alignment. However, Figure~\ref{fig:triplet_result}(b) shows general VLMs struggle with tissue recognition. CLIP achieves 59.1\% IV-matching but only 1.4\% IVT-matching, while LLaVA produces no IVT-matched triplets (0.0\%) despite 48.9\% IV-matching. Surgical VLMs demonstrate stronger tissue recognition but lower overall RAS scores. PeskaVLP leads with 20.8\% IVT-matching, followed by SurgVLP (11.9\%) and HecVL (7.6\%). 
%Combining IVT- and IT-matching for tissue-level recognition, surgical VLMs achieve 11.3--25.2\% versus general VLMs' 0.1--2.3\%.

\subsection{Causal Graph Analysis on LVLM}

Two examples of LLaVA's causal graph analysis on instrument classification are presented in Figure~\ref{fig:causal}. 
Even when selecting the top five heads most focused on visual tokens, the CLEANN algorithm reveals that many text tokens still emerge as direct causal contributors to the prediction. 
This suggests that certain keywords, such as \emph{surgical}, \emph{image}, and \emph{instrument}, may exert a direct influence to the final prediction than visual tokens. 
% even when evaluating only the attention heads with high focus on visual tokens, the results from applying CLEANN to derive the direct causal pathways show that LLaVA’s generation process does not fully rely on visual tokens. 
% Instead, many generated text tokens—such as \emph{surgical}, \emph{image}, and \emph{instrument}—contribute directly to the final prediction. 
Additionally, most visual tokens do not lie within clinically relevant regions for instrument classification. Instead, they are often scattered, with no consistent spatial pattern observed across examples. More causal graph examples are provided in the supplementary section E.3.

\subsection{Attention-based Occlusion Analysis} \label{sec:result_occlusion_study}
Figure~\ref{fig:conterfactual visual explaination} shows result on SurgVLP, after occluding image regions based on the attention heatmaps, 28.2\% of frames that contained IVT (instruments, verbs, targets) matches before occlusion no longer contained IVT matches after occlusion in top-5 high similarity score predictions, while remaining matched frames had average similarity scores drop by 9.87 for the IVT-matched prompt predictions. Figure~\ref{fig:conterfactual visual explaination} (a) shows an example where matches were lost despite occluding irrelevant regions, while Figure~\ref{fig:conterfactual visual explaination} (b) shows an example where others persisted even when instruments were fully blocked. Additional counterfactual explanations are in supplementary material section E.2.

\section{Discussion}
\paragraph{Surgical VLM requires stronger visual supervision.}
% The observations may not provide answers to questions \textbf{A} and \textbf{B} stated in section ~\ref{sec:prob_def} that align with typical intuition. 
General VLMs can roughly localize instruments but often fail to accurately recognize their types. In contrast, surgical VLMs trained on surgical video data tend to achieve moderately higher classification accuracy, but this improved performance does not result from better attention to clinically relevant cues; in fact, surgical VLMs often present worse alignment between model attention and meaningful image regions in our benchmarked tasks. For general VLMs, the low accuracy is primarily due to the lack of surgical domain knowledge. Surgical VLMs, despite being trained on domain-specific data, still fail to achieve satisfactory performance and often even weaker ability to attend to relevant cues. This may be because, 
% two key factors: (1) the visual similarity of the same organ or tissue in the background may unintentionally act as a shortcut for models' predictions, as supported by our occlusion experiments in Figure~\ref{fig:conterfactual visual explaination}.(2) 
in the surgical lecture videos, the language supervision in the video datasets is often provided at the video-clip level rather than the frame level. This weak supervision may encourage models to rely on high-level contextual patterns, such as visually similar organs or tissues in the background, rather than accurately localizing and recognizing individual instruments, as supported by our occlusion experiments in Figure~\ref{fig:conterfactual visual explaination}.

\textbf{LVLMs in Surgical applications may require supervision on reasoning.}
According to the causal graph analysis, current LVLM may rely more on text tokens to make decisions, at least on the tasks that are benchmarked in this work. 
% In the causal graph analysis, different attention heads focus on different information flows. 
% However, even when selecting the top five heads most focused on visual tokens, the CLEANN algorithm reveals that many text tokens still emerge as direct causal contributors to the prediction. This suggests that certain keywords in the question prompts or generated text may exert a stronger influence than visual tokens in guiding the model's output. 
For many surgical tasks, the text prompt may not provide deterministic cues, and decisions should be grounded more in visual information. Therefore, this behavior is suboptimal and suggests that model training should encourage more reliance on visual cues for decision-making.

\textbf{Limitations and Future work.} 
% As a first benchmark aiming to evaluate the explainability of VLMs in surgical tasks, we acknowledge that several aspects of our work present opportunities for future improvement and exploration. 
Our work focuses on instrument and action classification, although these are two essential tasks in the domain, other important tasks such as tissue or organ identification, phase recognition, and workflow analysis should be explored in the future. Although the proposed explainability-based benchmark pipeline can be flexibly extended to these tasks, defining clinically relevant cues and extracting them as ground truth for evaluating model attention and causal reasoning remain open challenges. Addressing these issues will require closer collaboration with clinical experts.
% In surgical action recognition, meaningful cues are conveyed not only by instruments but also by the surrounding deformable tissues and their motion, which may influence the model's decision-making. Therefore, although our RAFT-based action mask generation method can highlight most motion areas in the image, it can still be further improved, as large camera motions and redundant tissue movements may not be fully corrected by basic motion compensation. Additionally, future work should investigate effective prompting strategies that incorporate surgical domain knowledge to better understand VLM reasoning in clinical settings.

\section{Conclusion}
To bridge the gap in understanding the limitations of existing VLMs for surgical tasks, we present SurgXBench, the first benchmark that emphasizes explainability-driven analysis of VLMs in surgical domain.
% Explainability of VLMs in surgical tasks has received limited attention. In this work, we present the first benchmarks for evaluating VLM explainability in surgery.
We analyze the attention behavior and causal mechanisms of three general VLMs and three surgical VLMs
% , two general VLMs, and one LVLM 
both qualitatively through attention-based occlusion analysis and quantitatively by proposing novel explainability-based metrics. 
% Additionally, we introduce a RAFT-based approach for efficient annotation of instrument-tissue interaction regions to support explainability-based metrics in triplet recognition. 
This benchmark will serve as a foundation for advancing trustworthy multimodal models in surgical applications.

\bibliographystyle{plain}  
\bibliography{main}

\appendix
\newpage
\begin{center}
{\Large\bfseries Supplementary Material}
\end{center}
\vspace{0.5cm}
\section{Post-hoc XAI Visualization Techniques - Mathematical Formulations} \label{sec:attn_vis_supp}
This section provides the detailed mathematical formulations for the post-hoc XAI visualization methods described in Section 3.2 of the main paper.

\subsection{Grad-CAM for ResNet VLM}
For VLMs that use ResNet-style backbones~\cite{yuan2023surgvlp,yuan2024hecvl,yuan2024peskavlp}, the Grad-CAM formulation is defined as follows. Let $I \in \mathbb{R}^{H' \times W' \times 3}$ be the input image, and let $A \in \mathbb{R}^{K \times H \times W}$ denote the output of a selected convolutional layer in the image encoder, where $K$ is the number of channels and $H \times W$ is the spatial resolution. Let $A_k(i,j)$ denote the activation at spatial position $(i,j)$ in channel $k$. Let $\mathcal{P} = \{P_1, P_2, \dots, P_N\}$ be a set contain $N$ candidate natural language prompts. The vision-language model computes an embedding for the image $\phi_{\text{img}}(I)$ and for each text prompt $\phi_{\text{text}}(P_k)$, and calculates a similarity score $S_{P_k}$ (typically cosine similarity) for each prompt in $\mathcal{P}$:
\[
S_{P_k} = \frac{\phi_{\text{img}}(I)^\top \phi_{\text{text}}(P_k)}{\|\phi_{\text{img}}(I)\| \cdot \|\phi_{\text{text}}(P_k)\|}.
\]

The predicted prompt $\hat{P}$ is then selected as the one with the highest similarity score:
\[
\hat{P} = \arg\max_{P_k \in \mathcal{P}} S_{P_k},
\]
and let $S_{\hat{P}}$ denote the scalar similarity score corresponding to the predicted prompt $\hat{P}$. Grad-CAM computes the gradient of the score $S_c$ with respect to the feature map $A_k(i,j)$, and then applies global average pooling over spatial dimension:
\[
\alpha_k = \frac{1}{H \cdot W} \sum_{i=1}^H \sum_{j=1}^W \frac{\partial S_{\hat{P}}}{\partial A_k(i,j)}.
\]
The gradient tells how much each feature map channel $A_k$ contributes to the model's prediction $S_{\hat{c}}$. it does this by averaging the gradient of predicted score with respect to all locations $(i,j)$ in that channel and use it act as a weight that reflects the importance of that channel. Based on it, to compute the Grad-CAM heatmap, we first obtain a weighted combination of the activation maps using channel-wise importance weights $\alpha_k$, followed by a ReLU. Finally, we apply bilinearly upsampling to resize $L$ to the size of original image.

\begin{equation}\label{eq:gradcam-final}
L_{\text{Grad-CAM}} = \text{Upsample}_{\text{bilinear}}\left(\text{ReLU}\left( \sum_{k=1}^K \alpha_k \cdot A_k \right)\right).
\end{equation}

\subsection{Gradient-Based Attention Rollout for CLIP-ViT}
Following the method from \cite{chefer2021generic}, in transformer model, we leverage the self-attention matrices and their gradients to identify input image patches that contribute most to the similarity between an image and a given text prompt. Let $ \mathcal{T}^{(l)} \in \mathbb{R}^{N \times N} $ denote the self-attention matrix from the $ l $-th transformer block in the ViT encoder, where $ N $ is the number of tokens, a special token added at the beginning of the input, and its output representation is used to summarize the entire image for classification tasks. We extract $ \mathcal{T}^{(l)} $ after the softmax operation during the forward pass.

We compute the gradient of the similarity score with respect to the attention map:
\[
\nabla \mathcal{T}^{(l)} = \frac{\partial S_{\hat{P}}}{\partial \mathcal{T}^{(l)}},
\]
where $ S_{\hat{P}} $ is the predicted similarity score corresponding to the selected text prompt in a zero-shot setting. To capture influence on the similarity score, we compute the gradient-weighted attention map:
\[
\tilde{\mathcal{T}}^{(l)} = \text{ReLU} \left( \nabla \mathcal{T}^{(l)} \odot \mathcal{T}^{(l)} \right),
\]
where $ \odot $ denotes the elementwise (Hadamard) product. This highlights token-to-token interactions that both receive high attention and significantly impact the similarity score. To propagate relevance through the transformer layers, we recursively update a relevance matrix $ \mathbf{R}^{(l)} \in \mathbb{R}^{N \times N} $ as follows:
\[
\mathbf{R}^{(L)} = I, \quad \mathbf{R}^{(l)} = \mathbf{R}^{(l+1)} + \tilde{\mathcal{T}}^{(l)} \mathbf{R}^{(l+1)},
\]
starting from the final layer $ L $, where $ I $ is the $ N \times N $ identity matrix. This recursive formulation accumulates relevance flow across all layers. Let each patch token indexed by $k \in \{1,\dots,N-1\}$, after propagating to the input layer, we extract the relevance scores from the \texttt{[CLS]} token to each image patch token:
\[
r_k = \mathbf{R}^{(0)}_{\texttt{CLS},\, k}, \quad \text{for } k = 1, \dots, N-1,
\]
where $ r_k $ denotes the relevance of the $ k $-th image patch token to the final similarity score.

Finally, we reshape and upsample the vector $ [r_1, \dots, r_{N-1}] \in \mathbb{R}^{N-1} $ into the input image size using bilinear interpolation. The final heatmap can be constructed as following:
\begin{equation}\label{eq:Gradcam-vit}
L_{\text{Grad-CAM-ViT}} = \text{Upsample}_{\text{bilinear}}\left( \text{reshape}\left( [r_1, \dots, r_{N-1}] \right) \right),
\end{equation}

\subsection{Grad-CAM for Multimodal Large Language Models}\label{sec: gradcamllm}
For large language models, we adopt a method inspired by~\cite{zhang2024redundancy}. Since the predicted classes are generated as natural language, we apply Grad-CAM to specific tokens in the output sequence. To obtain a Grad-CAM heatmap for a given token \( t_j \), we isolate its corresponding logit \( z_j \) and compute the gradient with respect to the visual feature maps \( A_k \) from the last layer of the language model's decoder. Specifically, we compute:
\[
G_k = \frac{\partial z_j}{\partial A_k}.
\]
The remaining steps follow the standard Grad-CAM procedure described earlier.

\subsubsection{CLEANN: Causal Learning for Attention Networks}
Unlike contrastive learning, where taking the derivative of the similarity score with respect to visual tokens or feature maps can provide straightforward visual explainability, large vision-language models (LVLMs) involve more complex reasoning and interaction processes. As discussed in \cite{zhang2024redundancy}, Grad-CAM in LVLMs may indicate where the model focuses during generation, but it does not fully capture the underlying reasoning. As a result, our attention analysis procedure for LLaVA serves more as a soft interpretability measure. To better support our investigation, we adopt the Causal Graph Analysis method CLEANN \cite{rohekar2023causal,stan2024lvlm}, which provides a more principled view of the model's internal reasoning. For target token $t$ and relevant token set $\mathcal{T}$, the method tests conditional independence between token pairs $X, Y \in \mathcal{T}$ given conditioning set $\mathcal{Z} \subset \mathcal{T}$ using their attention distributions, where $\mathbf{a}_X$ represents the attention vector (row) of token $X$ in the attention matrix. This approach enables the discovery of how visual attention patterns causally influence language generation beyond simple correlation in multi-modal transformer architectures.
\section{Camera Motion Correction - Mathematical Formulation} \label{sec:camera_correction_math}

This section provides the detailed mathematical formulation for the camera motion correction method described in the main paper.

\subsection{Camera Motion Model}
We model the global camera motion as a linear combination of four basic operations: Pan ($P$), Tilt ($T$), Zoom ($Z$), and Roll ($R$). Each operation corresponds to a known optical flow prototype vector at any given pixel coordinate $(x, y)$ relative to the image center:

\textbf{Pan (p):} $p = \begin{pmatrix} -1 \\ 0 \end{pmatrix}$, 
\textbf{Tilt (t):} $t = \begin{pmatrix} 0 \\ -1 \end{pmatrix}$, 
\textbf{Zoom (z):} $z = \begin{pmatrix} -x \\ -y \end{pmatrix}$, 
\textbf{Roll (r):} $r = \begin{pmatrix} y \\ -x \end{pmatrix}$.

The predicted flow from the camera, $f_{\text{cam}}$, is a weighted sum of these prototypes: 
$$f_{\text{cam}} = P \cdot p + T \cdot t + Z \cdot z + R \cdot r$$

\subsection{Bidirectional Optimization}
To find the actual camera motion parameters robustly, we first use RAFT to compute total flow in both directions—the forward flow ($f_{\text{fwd}}$) and the backward flow ($f_{\text{bwd}}$)—and solve a least-squares problem over the valid pixel domain $\Omega \subset \mathbb{R}^2$. The optimization goal is to find the parameters $P, T, Z, R$ that minimize the total squared error $E = E_{\text{fwd}} + E_{\text{bwd}}$, which is the sum of the error contributions from both directions. 

The forward error component is:
\begin{align}
&E_{\text{fwd}} = \sum_{(x,y) \in \Omega} \| f_{\text{cam}}(x,y) - f_{\text{fwd}}(x,y) \|^2 \notag \\
               &= \sum_{(x,y) \in \Omega} \| (P \cdot p + T \cdot t + Z \cdot z + R \cdot r) - f_{\text{fwd}}(x,y) \|^2
\end{align}

The backward error component is:
$$E_{\text{bwd}} = \sum_{(x,y) \in \Omega} \| f_{\text{cam}}(x,y) + f_{\text{bwd}}(x,y) \|^2,$$

where the backward error assumes that camera motion should be consistent in both temporal directions, with the backward flow representing the negative of the forward camera motion. 

Solving this enhanced problem yields the best-fit parameters for the camera's movement. With these parameters, we construct the pure camera flow field, $f_{\text{cam}}$. Finally, we correct the original forward flow by subtracting this calculated camera motion, leaving only the motion of the independent objects: 
$$f_{\text{corrected}} = f_{\text{fwd}} - f_{\text{cam}}$$

\section{Implementation Details}
\subsection{Parameters}
All heatmaps from post-hoc XAI methods are thresholded using percentile-based Grad-CAM, where $\tau = 0.8$ selects the top 20\% high-attention regions. For RAFT-derived binary masks, we use a motion threshold of $\gamma = 0.8$. Across CLIP, BLIP, SurgVLP, HecVL, and PeskaVLP, a class is predicted if its similarity score exceeds the 90th percentile ($\delta = 0.9$). For the triplet recognition task, due to the large number of class combinations, we select the top 5 classes with highest similarity scores as predictions. CLEANN is implemented with key parameters: attention threshold $\kappa = 0.01$ to filter relevant tokens; p-value threshold $\alpha = 10^{-5}$ to enforce strong statistical significance in conditional independence tests; sample size $n = 128$ for reliable inference; and a cap of 50 image tokens per analysis to balance efficiency and coverage. Rather than analyzing all attention heads, we select the top-5 heads that prioritize visual over textual content. Head importance is computed as  
$
\text{head\_importance} = \frac{\max(\text{image\_attention})}{\max(\text{text\_attention})}
$
where higher values indicate greater visual focus for the target prediction.
\subsection{Prompts Design}\label{sec:prompt_design}
To align with VLMs' prompt-based input format, we construct template-based prompts. For instrument classification, we use: \textit{``What surgical tools do you see? Choose from: Grasper, Bipolar, Hook, Scissors, Clipper, Irrigator, Bag''} covering all instrument classes.  For triplet reasoning, we design a detailed prompt that guides the model to identify surgical actions by providing instrument capabilities and target options. The prompt instructs the model to look at the surgical image carefully and identify all visible actions, providing specific lists of what each instrument can do (e.g., grasper can grasp, retract, dissect, or manipulate tissue) and all possible surgical targets (cystic\_plate, gallbladder, omentum, etc.). We constrain the output to a three-word format: ``instrument verb target''.

\section{ Metrics}
\subsection{Standard Instrument Classification Metrics}  
Let TP, FP, and FN denote the number of true positives, false
positives, and false negatives across all instrument classes. The metrics are computed as:  
$\text{Precision} = \text{TP} / (\text{TP} + \text{FP})$,  
$\text{Recall} = \text{TP} / (\text{TP} + \text{FN})$,  
$F1 = 2 \cdot \text{Precision} \cdot \text{Recall} / (\text{Precision} + \text{Recall})$.

\subsection{Standard Triplet Recognition Metrics:}  
We adopt the standard triplet metrics from \cite{twinanda2022endonet}. Each predicted action is represented as a triplet of class labels $\langle \textit{instrument}, \textit{verb}, \textit{target} \rangle = (s, v, o)$, with the corresponding ground-truth triplet denoted as $(s^*, v^*, o^*)$. Binary match indicators are defined as:
\[
\text{Match}(s, v, o; s^*, v^*, o^*) = (\mathbbm{1}_{s = s^*},\ \mathbbm{1}_{v = v^*},\ \mathbbm{1}_{o = o^*}).
\]
The standard evaluation metrics focus on three specific matching conditions: IVT (instrument–verb–target), IV (instrument–verb), and IT (instrument–target). These metrics provide a finer understanding of the model’s ability to recognize structured surgical actions, which are inherently instrument-centric; the verb and target gain full semantic meaning only in the context of the acting instrument.
\section{Additional Results}
\subsection{Instrument classification}
\begin{figure*}[t!]
    \centering
    \includegraphics[width=\textwidth]{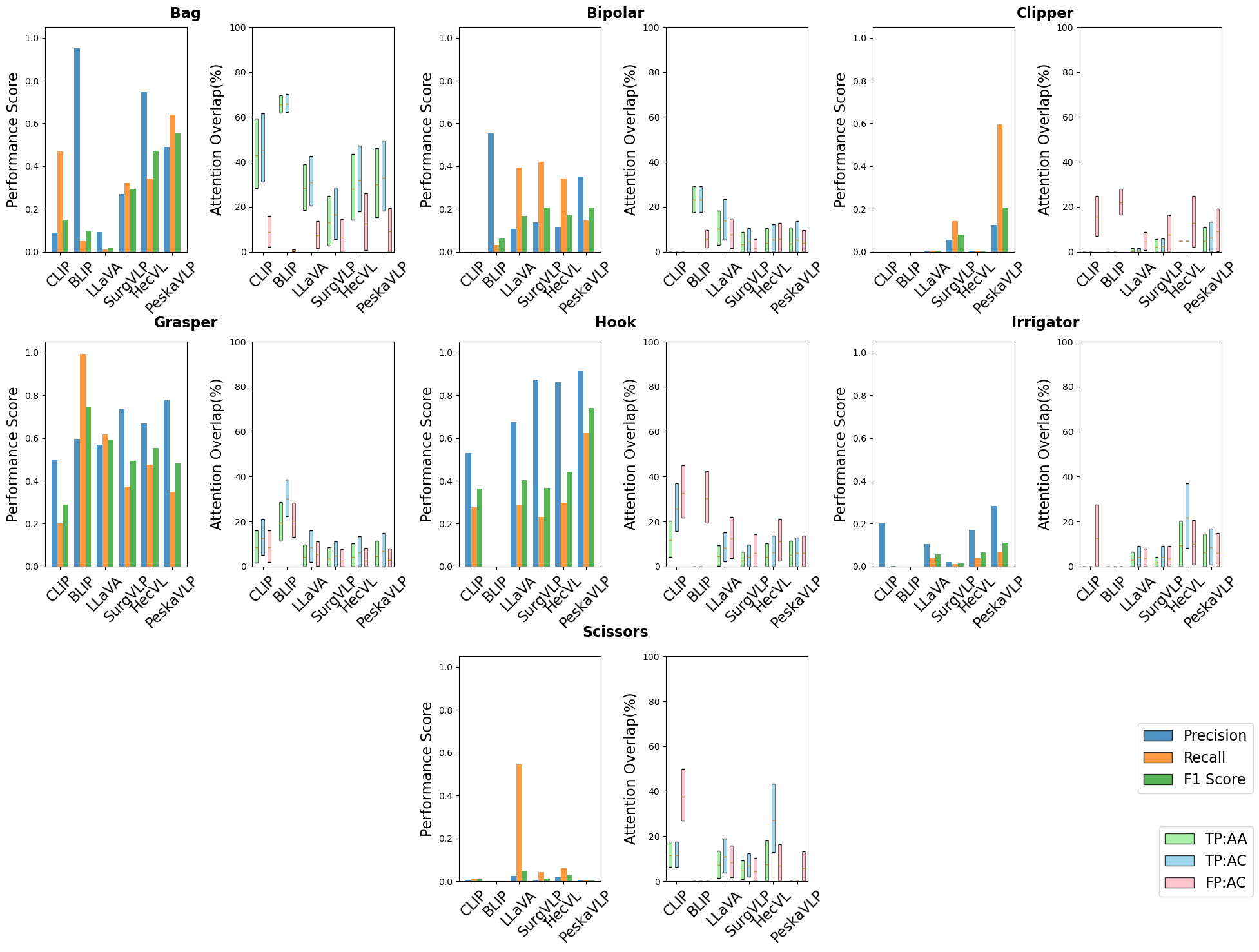}
        \caption{Surgical instrument detection performance and attention alignment across seven instrument types from the \textbf{Cholec80BBox} dataset. For each instrument, the left subplot shows classification performance (Precision, Recall, F1 Score) and the right subplot displays attention overlap AA and AC score between heatmaps and ground truth regions as box plots for True Positive and False Positive predictions.}\label{fig: AA_AC_boxresults}

\end{figure*}

Addition results include pericison, recall and box plot of attentiona quality metrics for instrument classification result are provided in figure~\ref{fig: AA_AC_boxresults}
\subsection{Occlusion Analysis}
\begin{figure*}[!htbp]
    \centering
    \includegraphics[width=\linewidth]{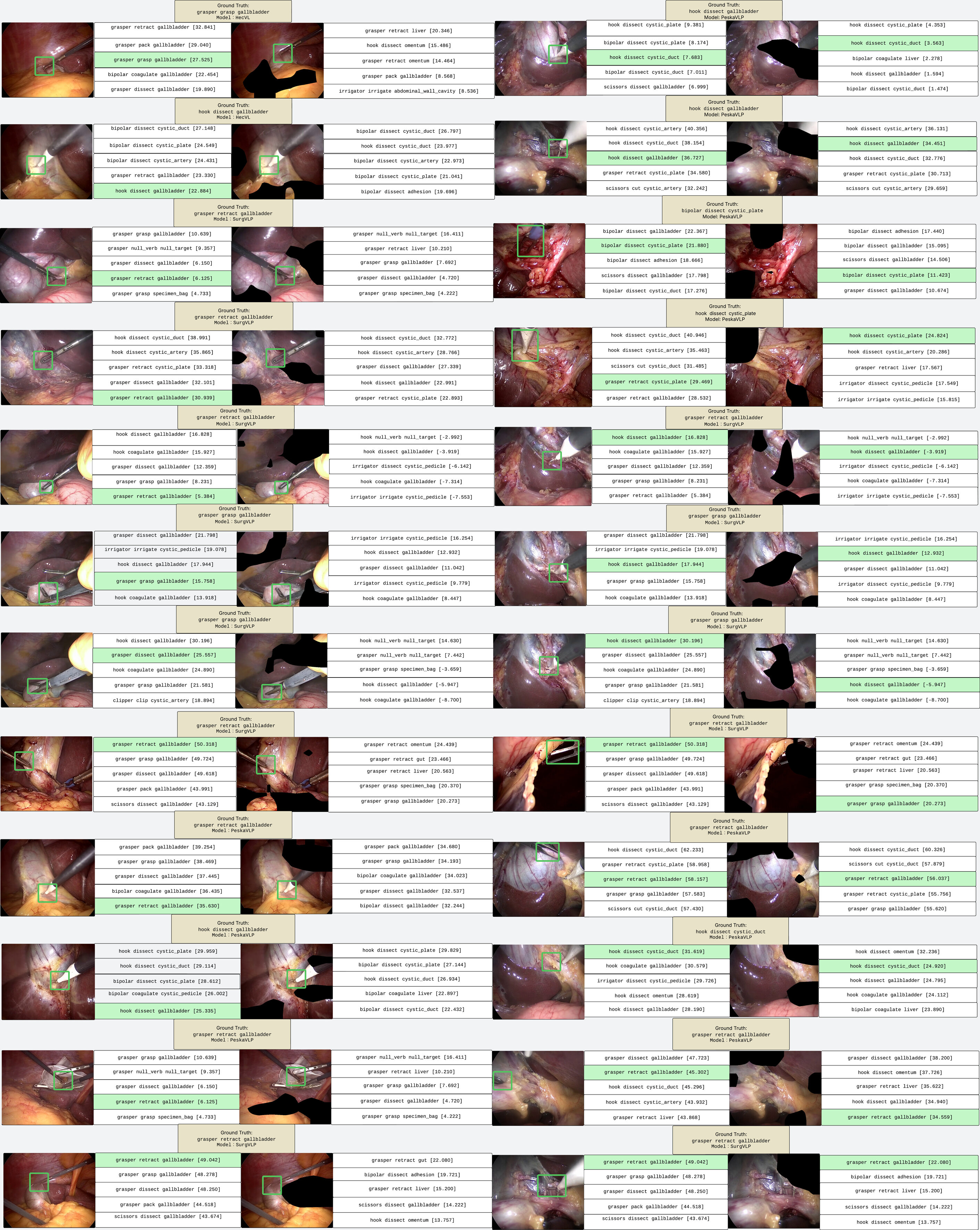}
    \caption{Occlusion analysis results for SurgVLP, HecVLP, and PeskaVLP. }
    \label{fig:occlusion_all}
\end{figure*}

Additional results for the occlusion analysis are provided in Figure~\ref{fig:occlusion_all} showing successful tool alignment under occluded conditions. The left panel shows cases where occlusion blocks irrelevant areas, yet the correct IVT(instrument-verb-target) match is no longer among the top-5 predictions. The right panel shows cases where relevant regions are occluded, but the correct prediction still appears in the top-5. Prompts highlighted in green indicate triplets that achieve IVT coverage with the ground truth, while gray prompts represent cases where the instrument was not correctly identified. The green bounding box on the images represents the instrument-relevant area in the triplets.

\subsection{Causal Graph Analysis}
\begin{figure*}[t]
    \centering
    \includegraphics[width=0.9\linewidth]{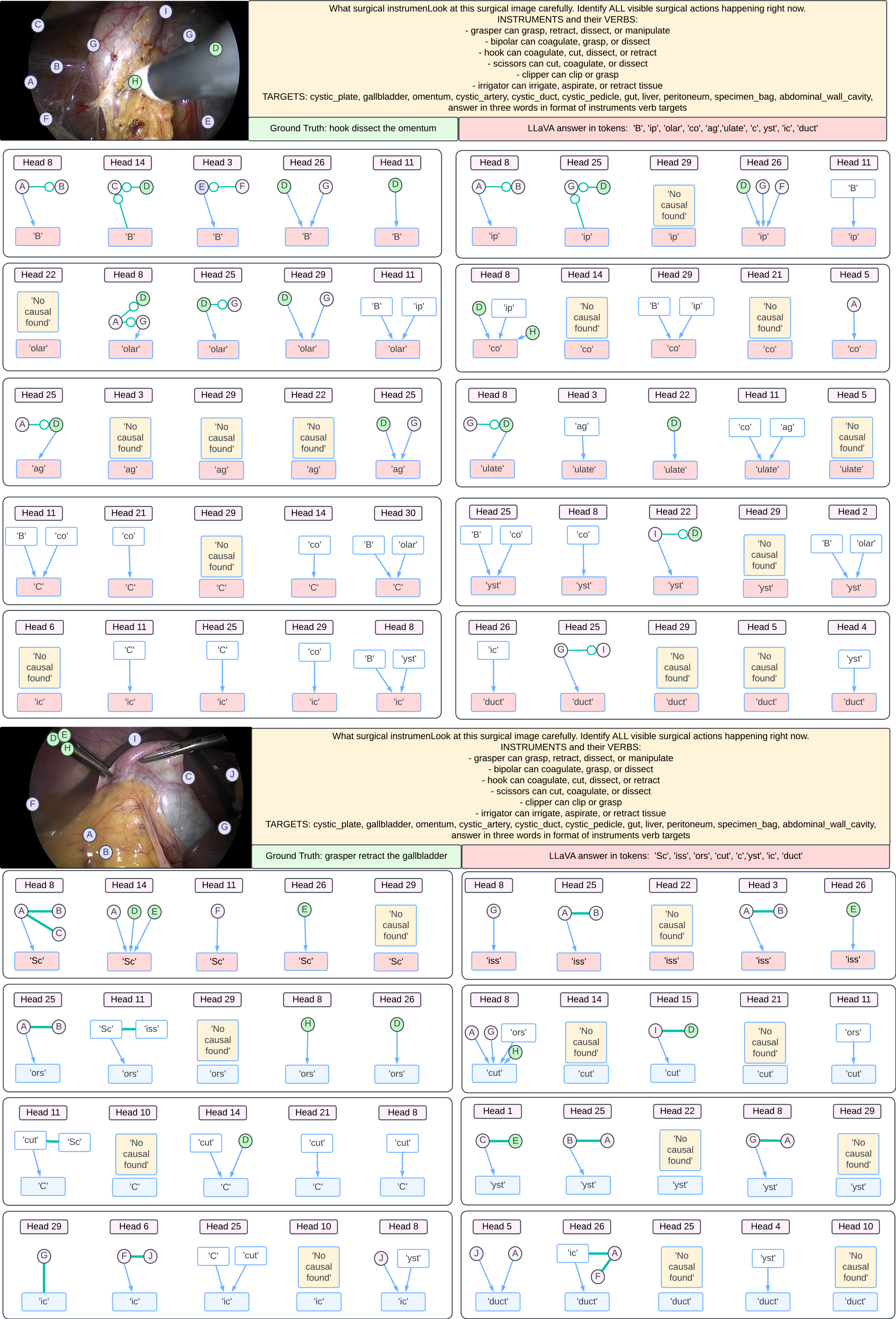}
\caption{This figure shows triplet recognition examples from LLaVA}\label{fig:causal1}
\end{figure*}

Two additional results for triplet recognition are shown in Figure~\ref{fig:causal1}. These figures highlight same findings discussed in Section~5 of the main paper, causal graphs generated using the CLEANN method (radius 3, top-5 visually important attention heads). Generated responses are in token format where each box contains the causal graph of each head for each token. Green circles indicate visual tokens that are in or close to the relevant area, and white boxes represent text tokens that were previously generated.


%\begin{figure*}[t]
%    \centering
%    \includegraphics[width=0.9\linewidth]{causal_example2.jpeg}
 %   \caption{Example 2 of causal graph analysis showing similar attention spread without clear focus.}
 %   \label{fig:causal2}
%\end{figure*}
\subsection{Text token ratio Analysis}
Text token ratio analysis using two complementary methods: (i) a CLEANN-based causal selection that identifies tokens directly influencing target generation through conditional independence testing and computes text-visual ratios among causal dependencies, and (ii) an attention-based approach that calculates text-visual ratios by summing attention weights across text and image regions. Both approaches detected substantial text over-reliance: attention analysis showed 91-92 \% text dominance for tokens like "yst", "plate", and "cut", while CLEANN revealed genuine text bias in tokens such as "cutting" (77.6 \% text dominance). The consistent detection of high text dominance across both methodologies confirms that surgical vision-language models exhibit over-reliance on linguistic context rather than visual evidence.
\begin{table}[htbp]
\centering
\caption{Comparison of Attention-Based vs. Causal Analysis for Text Dominance Detection}
\label{tab:method_comparison}
\begin{tabular}{llcc}
\toprule
\textbf{Subword Token} & \textbf{Method} & \textbf{Text Dom.} & \textbf{Visual Dom.} \\
\midrule
\multirow{2}{*}{\texttt{cut}} 
& Attention & 0.922 & 0.078 \\
& CLEANN & 0.328 & 0.672 \\
\midrule
\multirow{2}{*}{\texttt{yst}\textsuperscript{\dag}} 
& Attention & 0.920 & 0.080 \\
& CLEANN & 0.133 & 0.967 \\
\midrule
\multirow{2}{*}{\texttt{plate}} 
& Attention & 0.918 & 0.082 \\
& CLEANN & 0.240 & 0.946 \\
\midrule
\multirow{2}{*}{\texttt{re}\textsuperscript{\dag}} 
& Attention & 0.924 & 0.076 \\
& CLEANN & 0.195 & 0.805 \\
\midrule
\multirow{2}{*}{\texttt{cutting}} 
& Attention & 0.913 & 0.087 \\
& CLEANN & 0.776 & 0.224 \\
\midrule
\multirow{2}{*}{\texttt{gr}\textsuperscript{\dag}} 
& Attention & 0.917 & 0.083 \\
& CLEANN & 0.000 & 1.000 \\
\bottomrule
\end{tabular}
\begin{flushleft}
\textsuperscript{\dag} Subword fragment from tokenization (e.g., \texttt{yst} from ``cystic'', \texttt{re} from ``retract'', \texttt{gr} from ``grasper'')
\end{flushleft}
\end{table}
\end{document}